\def\year{2020}\relax
\documentclass[letterpaper]{article} 
\usepackage{aaai20}  
\usepackage{times}  
\usepackage{helvet} 
\usepackage{courier}  
\usepackage[hyphens]{url}  
\usepackage{graphicx} 
\usepackage{graphicx}  
\usepackage{booktabs}
\usepackage{amssymb}

\newcommand{\dataset}{JEC-QA}

\newcommand{\typea}{KD-questions}
\newcommand{\typeb}{CA-questions}

\newcommand{\questionn}{26,365}
\newcommand{\readingn}{79,433}

\newcommand{\support}[1]{\emph{#1}}
\newcommand{\citet}[1]{\citeauthor{#1} \shortcite{#1}}
\newcommand{\citep}{\cite}

\title{\dataset: A Legal-Domain Question Answering Dataset}
\author{
Haoxi Zhong,$^\ast$\textsuperscript{\rm 1}
Chaojun Xiao,\thanks{Indicates equal contribution.}\textsuperscript{\rm 1}
Cunchao Tu,\textsuperscript{\rm 1} 
Tianyang Zhang,\textsuperscript{\rm 2}
Zhiyuan Liu\thanks{Corresponding author.}\textsuperscript{\rm 1},
Maosong Sun\textsuperscript{\rm 1} \\\\ 
\textsuperscript{\rm 1}Department of Computer Science and Technology\\
Institute for Artificial Intelligence, Tsinghua University, Beijing, China\\
Beijing National Research Center for Information Science and Technology, China
 \\
\textsuperscript{\rm 2}Beijing Powerlaw Intelligent Technology Co., Ltd., China\\
zhonghaoxi@yeah.net, \{xcjthu,tucunchao\}@gmail.com, zty@powerlaw.ai, \{lzy,sms\}@tsinghua.edu.cn
}

\begin{document}

\maketitle

\begin{abstract}
We present \dataset, the largest question answering dataset in the legal domain, collected from the National Judicial Examination of China. The examination is a comprehensive evaluation of professional skills for legal practitioners. College students are required to pass the examination to be certified as a lawyer or a judge. The dataset is challenging for existing question answering methods, because both retrieving relevant materials and answering questions require the ability of logic reasoning. Due to the high demand of multiple reasoning abilities to answer legal questions, the state-of-the-art models can only achieve about $28\%$ accuracy on \dataset, while skilled humans and unskilled humans can reach $81\%$ and $64\%$ accuracy respectively, which indicates a huge gap between humans and machines on this task. We will release \dataset\ and our baselines to help improve the reasoning ability of machine comprehension models. You can access the dataset from \url{http://jecqa.thunlp.org/}.
\end{abstract}

\section{Introduction}
\label{sec:intro}
\textbf{L}egal \textbf{Q}uestion \textbf{A}nswering (\textbf{LQA}) aims to provide explanations, advice or solutions for legal issues. A qualified LQA system can not only provide a professional consulting service for unskilled humans but also help professionals to improve work efficiency and analyze real cases more accurately, which makes LQA an important NLP application in the legal domain. Recently, many researchers attempt to build LQA systems with machine learning techniques~\cite{fawei2018methodology} and neural network~\cite{do2017legal}. Despite these efforts in employing advanced NLP models, LQA is still confronted with the following two major challenges. The first is that there is less qualified LQA dataset which limits the research. The second is that the cases and questions in the legal domain are very complex and rigorous. As shown in Table~\ref{table:exampleA}, most questions in LQA can be divided into two typical types: the knowledge-driven questions (\typea) and case-analysis questions (\typeb). \typea\ focus on the understanding of specific legal concepts, while \typeb\ concentrate more on the analysis of real cases. Both types of questions require sophisticated reasoning ability and text comprehension ability, which makes LQA a hard task in NLP.
 
\begin{table}[t]
\centering
\resizebox{.95\columnwidth}{!}{
\begin{tabular}{p{0.925\columnwidth}}
\toprule
\textbf{Knowledge-Driven Question:} 
Which of the following belong to the ``property'' of Civil Law? \\ 
\textbf{Option:}\\
$\times$ A. Trademark.  \quad $\times$ B. The star on the sky. \\
$\times$ C. Gold teeth. \quad $\checkmark$ D. Fish in the pond. \\
\midrule
\textbf{Case-Analysis Question:} 
Alice owed Bob 3,000 yuan. Alice proposed to pay back with 10,000 yuan of counterfeit money. Bob agreed and accepted it. Which crimes did Alice commit? \\ 
\textbf{Option:}\\ 
$\checkmark$ A. Crime of selling counterfeit money.\\
$\times$ B. Crime of using counterfeit money.\\
$\times$ C. Crime of embezzlement. \\
$\times$ D. Alice did not constitute a crime.\\
\bottomrule
\end{tabular}
}
\caption{Two typical examples of \typea\ and \typeb\ in LQA. All examples we show in the paper are translated from Chinese for illustration.}
\label{table:exampleA}
\end{table}

To push forward the development of LQA, we present \dataset\ in this paper, the largest and more challenging LQA dataset. \dataset\ collects questions from the National Judicial Examination of China (NJEC) and websites for the examination. NJEC is the legal professional certification examination for those who want to be a lawyer or a judge in China. Every year, only around $10\%$ of participants can pass the exam, proving it difficult even for skilled humans.

There are three main properties of \dataset:
(1) \dataset\ contains \questionn\ multiple-choice questions in total, with four options for each question. The number of questions in \dataset\ is 50 times larger than the previous largest LQA dataset~\cite{kim2016coliee}.
(2) \dataset\ provides a database including all the legal knowledge required by the examination. The database is collected from the National Unified Legal Professional Qualification Examination Counseling Book and Chinese legal provisions.
(3) \dataset\ provides extra labels for questions, including the type of questions (\typea\ or \typeb) and the reasoning abilities required by the questions. The meta information labeled by skilled humans will be useful for depth-analysis of LQA.

\dataset\ can be addressed following the setting of  OpenQA~\cite{chen2017reading,wang2017r,wang2017evidence,lin2018denoising}. That is, we need to retrieve relevant articles from the databases and apply reading comprehension models to answer questions. Distinct from existing question answering datasets~\cite{yang2015wikiqa,mctestdataset,cnndailymaildataset,squaddataset,newsqa2016,lai2017race}, \dataset\ requires multiple reasoning abilities to answer the questions including word matching, concept understanding, numerical analysis, multi-paragraph reading, and multi-hop reasoning. The detailed analysis can be found in the section of Reasoning Types.

\begin{figure}[t]
\centering
    \includegraphics[width=.95\columnwidth]{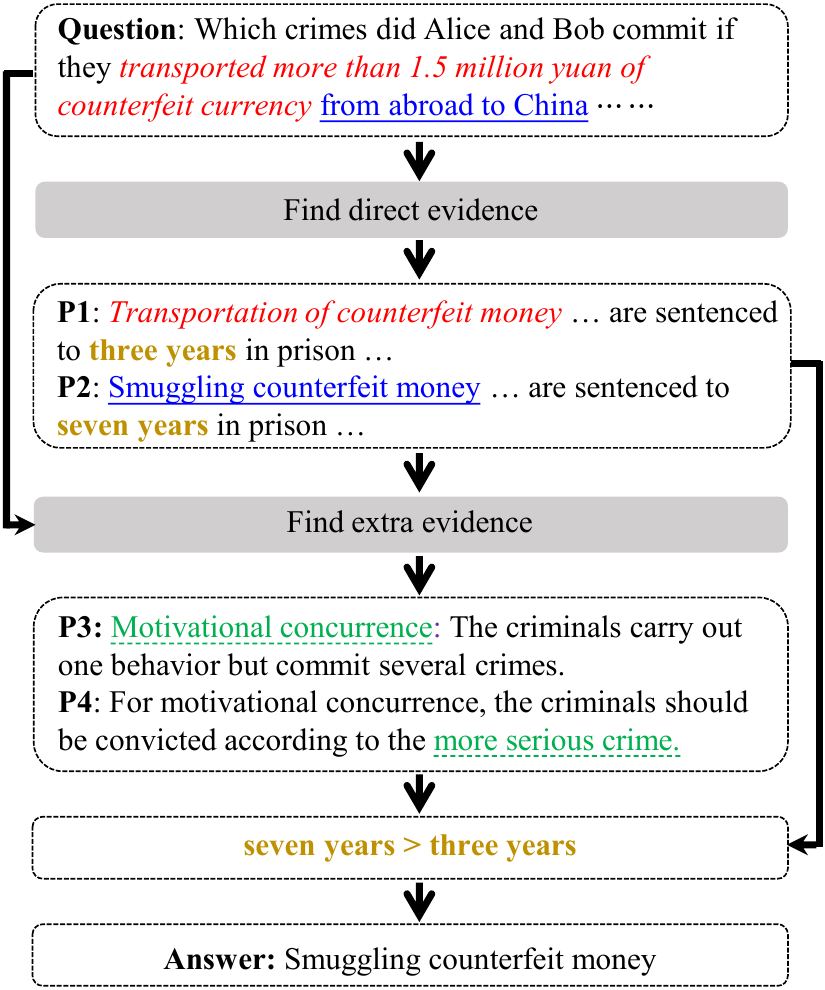}
\caption{An illustration of the logic that a person answers a question in \dataset. \textbf{P1} to \textbf{P4} are $4$ relevant paragraphs retrieved from the legal database. The first two are definitions of two crimes. The last two describe a legal concept and sentencing criterion. }
\label{fig:flowsheet}
\end{figure}

To get a better understanding of these reasoning abilities, we show a question of \dataset\ in Fig.~\ref{fig:flowsheet} describing a criminal behavior which results in two crimes. The models must understand ``Motivational Concurrence'' to reason out extra evidence rather than lexical-level semantic matching. Moreover, the models must have the ability of multi-paragraph reading and multi-hop reasoning to combine the direct evidence and the extra evidence to answer the question, while numerical analysis is also necessary for comparing which crime is more serious. We can see that answering one question will need multiple reasoning abilities in both retrieving and answering, makes \dataset\ a challenging task.

To investigate the challenges and characteristics of LQA, we design a unified OpenQA framework and implement seven representative neural methods of reading comprehension. By evaluating the performance of these methods on JEC-QA, we show that even the best method can only achieve about $25\%$ and $29\%$ on \typea\ and \typeb\ respectively, while skilled humans and unskilled humans can reach $81\%$ and $64\%$ accuracies on \dataset. The experimental results show that existing OpenQA methods suffer from the inability of complex reasoning on \dataset\ as they cannot well understand legal concepts and handle multi-hop reasoning.

In summary, \dataset\ is the largest LQA dataset, and it is more challenging compared with existing datasets due to the requirements of multiple reasoning abilities and legal knowledge. \dataset\ will benefit the research of question answering and legal analysis. We also show the performance of existing methods, conduct an in-depth analysis of \dataset\ and outlook the future research direction. You can access the dataset from \url{http://jecqa.thunlp.org/}.

\section{Related Work}

\label{sec:related}

\subsection{Reading Comprehension}

There have been numerous reading comprehension datasets proposed in recent years, such as  CNN/DailyMail~\cite{cnndailymaildataset}, MCTest~\cite{mctestdataset}, SQuAD~\cite{squaddataset}, WikiQA~\cite{yang2015wikiqa} and NewsQA~\cite{newsqa2016}. Deep reading comprehension models~\cite{seo2016bidirectional,wang2017gated,wang2016RC,dhingra2016gated,yih2015semantic} have achieved promising results on these early datasets. Besides, recent works like TrivialQA~\cite{triviaqa2017}, MS-MARCO~\cite{msmarco2016} and DuReader~\cite{dureader2017he} contain multiple passages for each question, while RACE~\cite{lai2017race}, HotpotQA~\cite{yang2018hotpotqa} and ARC~\cite{clark2018think} datasets require the ability of reasoning. Based on these datasets, researchers~\cite{wang2018co,wang2017r,wang2018multi,clark2017simple} propose to aggregate information from all passages. These datasets take a step towards a more challenging reading comprehension task, but still have a limitation that the answers can be extracted from the passages directly with semantic matching. As a result, existing RC systems are still lack of reasoning ability and language understanding~\cite{jia2017adversarial}.

\subsection{Open-domain Question Answering}
OpenQA is first proposed by \cite{green1961baseball}, which aims to answer questions with external knowledge bases, such as collected documents~\cite{voorhees1999trec}, web-pages~\cite{kwok2001scaling,chen2017discriminative} or structured knowledge bases~\cite{berant2013semantic,bordes2015large,yu2017improved}.

Most OpenQA models contain two steps: reading material retrieval and answer extraction/selection~\cite{chen2017reading,dhingra2016gated,cui2016attention}. Without document-level annotations, they retrieve documents with unsupervised information retrieval methods, e.g., TF-IDF or BM25 retriever. However, these models focus on the lexical similarity between articles and questions rather than semantic relevance. Recent approaches~\cite{lin2018denoising,wang2017evidence,clark2017simple} tend to rerank passages retrieved in the first step and filter out noisy contents. Although these methods can surpass human performance in certain situations, they are still lack of reasoning ability~\cite{rajpurkar2018know}.

\subsection{Legal Intelligence}
Owing to the massive quantity of high-quality textual data in the legal domain, employing NLP techniques to solve legal intelligence problems has been more and more popular in recent years, e.g., generating court views to interpret charge results~\cite{ye2018interpretable}, retrieving relevant or similar cases~\cite{chen2013text,raghav2016analyzing}, predicting charges or identifying applicable articles~\cite{luo2017learning,hu2018few,he2018secaps,zhong2018legal,xiao2018cail,shen2018legal}.

Meanwhile, answering legal questions has been a long-standing challenge for applications of legal intelligence.~\citet{kim2016coliee,kim2018coliee} held a legal question answering competition, where rule-based systems~\cite{fawei2018methodology} and neural models~\cite {do2017legal} were applied to this task.

In spite of this, we are still far away from applicable LQA systems, due to the poor performance, reasoning ability, and interpretability. We collect \dataset\ from NJEC, which can serve as a good benchmark of the reasoning ability of legal domain question answering models. 

\section{Dataset Construction and Analysis}

\subsection{Dataset Construction}

\textbf{Questions.} We collect $2,700$ multiple-choice questions from the $2009$ to $2017$ national judicial and $30,371$ practice exercises from websites. After removing duplicated questions, there are \questionn\ questions in \dataset.

\begin{table}[htb]
\centering
\begin{tabular}{c|c|c|c}
\toprule
           & \typea & \typeb & Total \\ \midrule
Single     & $4,603$                  & $8,738$                 & $13,341$ \\
Multi & $5,158$                  & $7,866$                  & $13,024$ \\\midrule
All      & $9,761$                  & $16,604$                 & $26,365$
\\ \bottomrule
\end{tabular}
\caption{The statistics of question types in \dataset.}
\label{table:count}
\end{table}

Each question in \dataset\ contains a question description and four candidate options. There are single-answer and multi-answer questions in \dataset. Meanwhile, we can also classify the questions into Knowledge-Driven Questions~(\typea) and Case-Analysis Questions~(\typeb). \typea\ pay attention to the definition and interpretation of legal concepts, while \typeb\ require analysis for the actual scenarios. Answering both types of questions requires reasoning ability. More detailed statistics of question types are summarized in Table~\ref{table:count}.

\begin{table}[htb]
\centering
\resizebox{.95\columnwidth}{!}{
\begin{tabular}{c|c|c|c}
\toprule
                 & Questions                 & Options & Paragraphs                 \\ \midrule
Count            & \questionn & 105,460 & \readingn \\  
Average Length      & $47.01$                     & $14.52$   & $58.42$                    \\
Max Length          & $547$                       & $153$     & $2,738$                    \\
Vocab Size       & $29,268$                    & $29,987$  & $47,808$                   \\ \midrule
Total Vocab Size & \multicolumn{3}{c}{70,110}                                    
\\ \bottomrule
\end{tabular}
}
\caption{The statistics of questions, options, and reading paragraphs in \dataset.}
\label{table:statistics}
\end{table}

\begin{table*}[!t]
\centering
\resizebox{2.0\columnwidth}{!}{
\begin{tabular}{p{3.5cm}lllp{11cm}}

\toprule
Reasoning Type     & KD-Q & CA-Q & All & Examples
\\

\midrule
Word Matching       & $\mathbf{65.9}\%$ & $23.9\%$ & $40.5\%$ & 
\textbf{Question}: Which option is a form of state compensation? \newline 
\textbf{Option}: \support{Monetary awards} \newline
\textbf{Paragraph}: \support{Monetary awards} is a form of state compensation. 
\\

\midrule
Concept Understanding            & $36.4\%$ & $42.8\%$ & $40.2\%$ &
\textbf{Question}: Who is the \support{principal offender} according to Criminal Law? \newline
\textbf{Option}: Bob, the leader of a robbery group, who ordered his subordinates to commit robbery on multiple occasions, but was never personally involved. \newline
\textbf{Paragraph}: The \support{principal offender} is the person in a group of offenders who leads, organizes, and carries out the main part of a criminal act.
\\

\midrule
Numerical Analysis    & $4.6\%$ & $14.9\%$ & $10.8\%$ &
\textbf{Question}: In which of the following circumstances should an extraordinary general meeting of shareholders be convened? \newline
\textbf{Option}: The registered capital of the company is \support{12 million yuan}, and the unrecovered loss is \support{5 million}. \newline
\textbf{Paragraph}: In the following circumstances, an extraordinary general meeting of shareholders should be convened: (1) When the unrecovered losses amount to \support{one-third of the total paid-up share capital}; ...
\\

\midrule
Multi-Paragraph Reading    & $19.7\%$ & $29.4\%$ & $25.5\%$ &
\textbf{Question}: Which statement is true about corporate crimes? \newline
\textbf{Option}: Corporates can be the subject of bank fraud.\newline
\textbf{Paragraph $1$}: Article $200$ of Criminal Law: The punishment of fraud offenses committed by corporates. If a corporate commits any crimes specified in \support{articles 192, 194, or 195 of this section}, it shall be fined. \newline
\textbf{Paragraph $2$}: Article $194$ of Criminal Law: \support{Bank fraud}...
\\

\midrule
Multi-Hop Reasoning & $8.33\%$ & $\mathbf{66.2}\%$ & $43.2\%$ & Shown in Fig.~\ref{fig:flowsheet}.\\

\bottomrule
\end{tabular}
}
\caption{Percentages and examples of questions in \dataset\ that require different types of reasoning. We only list one correct option in the table. One question may require multiple reasoning abilities so the sum of percentages is over $100\%$.}
\label{table:examples}
\end{table*}

\textbf{Database.} As mentioned in introduction, all necessary knowledge for the examination is involved in the National Unified Legal Professional Qualification Examination Counseling Book and Chinese legal provisions. The book contains $15$ topics and $215$ chapters with highly hierarchically formed contents. To guarantee the retrieval quality, we convert this papery book into structured electronic edition manually instead of using OCR (Optical Character Recognition) tools. For Chinese legal provisions,  we include $3,382$ different legal provisions in our database. The details of the database can be found in Table~\ref{table:statistics}.

\subsection{Reasoning Types}

We summarize $5$ different reasoning types required for answering questions in \dataset\ from \dataset\ and previous works~\cite{lai2017race,clark2018think}, and the examples are shown in Table~\ref{table:examples}.

\label{sec:rt}
(1) \textbf{Word Matching.} This is the simplest type of reasoning. The models only need to check which options are matched with the relevant paragraphs and the relevant paragraphs can be easily retrieved by simple search strategies as the contexts are highly consistent. Questions that require this type of reasoning are similar to the ones in traditional reading comprehension datasets.

(2) \textbf{Concept Understanding.} As our dataset is built on the legal domain, models need to understand legal concepts to answer these questions. As shown in the $2$-nd example in Table~\ref{table:examples}, models need to understand the meanings of ``principal offender'' to choose the correct answer.
    
(3) \textbf{Numerical Analysis.} This type of reasoning requires models to perform arithmetic operations. As shown in the $3$-rd example in Table~\ref{table:examples}, models must calculate $12\times\frac{1}{3}=4<5$ to answer it.

(4) \textbf{Multi-Paragraph Reading.} The settings for previous single-paragraph reading tasks guarantee that enough evidence can be found within one paragraph. However, as shown in the $4$-th example in Table~\ref{table:examples}, specific questions in \dataset\ require reading multiple paragraphs to gather enough evidence, which makes \dataset\ more challenging compared to traditional reading comprehension tasks.

(5) \textbf{Multi-Hop Reasoning.} Multi-hop reasoning means that we need multiple steps of logical reasoning to get the answers. Multi-hop reasoning is common in our real lives, but it is hard for existing methods to provide an interpretable reasoning process. Here we show an example of multi-hop reasoning in Fig.~\ref{fig:flowsheet}. Answering this question need to make several steps of reasoning, including concept understanding, numerical analysis, and multi-paragraph reading. From Table~\ref{table:examples}, we observe that more than $66\%$ \typeb\ require multi-hop reasoning ability, which leads great challenges to existing reading comprehension models.

In conclusion, we summarize that all $5$ types of reasoning above are essential for answering questions in \dataset\ and models need to handle these reasoning issues to achieve a promising performance in \dataset.

\section{Experiments}
In this section, we conduct detailed experiments and analysis to investigate the performance of existing question answering models on \dataset. Following the settings of OpenQA, we first retrieve relevant paragraphs and then employ question answering models to give answers.

\subsection{Retrieve Strategy}
To retrieve relevant materials from the database, we apply ElasticSearch\footnote{\url{https://www.elastic.co/}} to build a search engine containing the whole database. As the text materials are hierarchically structured, we store the contents into the search engine with meta-information, such as tags, chapter titles, and section titles. Because different options may focus on various aspects even within the same question, we need to retrieve reading paragraphs for each option separately. 

To reduce noisy data and narrow the scope during retrieving, we need to identify the topic (e.g., constitution, criminal law) of the questions. There are $15$ topics in total, and we employ $3$ representative models, including  BERT~\cite{devlin2018bert}, TextCNN~\cite{kim2014convolutional}, and DPCNN~\cite{johnson2017deep}. From $10,008$ labeled instances, we randomly select $1,956$ instances for testing and the rest for training. The performance of topic classification is shown in Table~\ref{table:topic}.  

\begin{table}[htb]
\centering
\begin{tabular}{c|c|c|c}
\toprule
      Method   & Top-$1$ & Top-$2$ & Top-$3$ \\ \midrule
TextCNN & $77.97$ & $87.14$ & $91.46$ \\
DPCNN & $75.16$ & $87.40$ &  $92.71$ \\\midrule
BERT     & $75.31$ & $88.60$ & $\mathbf{93.10}$ \\
\bottomrule
\end{tabular}
\caption{Accuracy (\%) of topic classification.}
\label{table:topic}
\end{table}

From the experimental results, we can see that the top-$1$ accuracy of topic classification is unsatisfactory, and the increment is little from top-$2$ to top-$3$ (only about $5\%$). In order to reach a balance of performance and speed, we employ BERT as our topic classifier to select the top-$2$ relevant topics and retrieve $K$ most relevant reading paragraphs for each topic. Besides, we also retrieve $K$ extra reading paragraphs from Chinese legal provisions. In total, we retrieve $3K$ paragraphs for each option. We choose $K=6$ for experiments and we will discuss the reason in Comparative Analysis.

To evaluate the performance of our retrieve strategy, we randomly select $377$ questions as Annotation Set for manual annotation and annotate each question with $3$ labels, including (1) All Hit (AH): all relevant paragraphs are successfully fetched. (2) Partial Miss (PM): some relevant paragraphs are missing. (3) All Miss (AM): no relevant paragraphs exist in the fetched results.

\begin{table}[htb]
\centering
\resizebox{.95\columnwidth}{!}{
\begin{tabular}{c|ccc}
\toprule
               Type         & AH & PM & AM \\ \midrule
All questions           & $45.69$	     & $35.77$	          & $18.54$    \\
\typea                  & $\underline{59.55}$        & $28.09$            & $12.36$    \\
\typeb                  & $38.76$        & $\underline{39.61}$            & $\underline{21.63}$    \\\midrule
Word Matching          & $\underline{62.22}$        & $26.67$            & $11.11$    \\
Concept Understanding   & $42.54$        & $35.82$            & $21.64$    \\
Numerical Analysis      & $38.89$        & $33.33$            & $\underline{27.78}$    \\
Multi-Paragraph Reading & $38.82$        & $\underline{48.24}$            & $12.94$    \\
Multi-Hop Reasoning     & $38.89$        & $37.50$            & $23.61$  
 \\
\bottomrule
\end{tabular}
}
\caption{Evaluation results (\%) of the retrieval strategy.}
\label{table:missing}
\end{table}  

The evaluation results are listed in Table~\ref{table:missing}. From this table, we observe that around $46\%$ of the questions can be answered correctly based on retrieved materials. The hit rate of \typea\ is significantly higher than \typeb\ as \typea\ are usually related to specific concepts, which leads to easier retrieval. Among different types of reasoning, the performance in word-matching questions achieves the highest hit rate of $62\%$ as the questions are highly consistent with reading paragraphs. The hit rates for other types achieve substantially lower scores due to the demand for sophisticated reasoning ability.

\subsection{Experiment Settings}

We employ a controlled experimental setting to ensure a fair comparison among various question answering models. Moreover, we use fastText~\cite{joulin2016fasttext} to pretrain word embeddings on a large-scale legal domain dataset~\cite{xiao2018cail}. For all models, the dimension of word embeddings is $w=200$ and the hidden size of model layers is $d=256$.

As the original tasks of our baselines are various, we design a unified OpenQA framework for them. More specifically, the input for the framework is a triplet $(q,o,r)$ representing the question, options, and reading paragraphs fetched in the retrieving step. $q$ is a sequence of words $(q_1,q_2,\ldots,q_{|q|})$. $o$ is a tuple of $n=4$ word sequences expressed as $((o_{1,1},o_{1,2},\ldots,o_{1,|o_1|}),\ldots,(o_{n,1},\ldots,o_{n,|o_n|}))$, corresponding to $n$ options. Suppose there are $m=18$ reading paragraphs for each option, then $r_{i,j}$ denotes the $j$-th reading paragraph for the $i$-th option, i.e., $r_{i,j}=(r_{i,j,1},r_{i,j,2},\ldots,r_{i,j,|r_{i,j}|})$, where $i \in [1, n]$ and $j \in [1, m]$.

For the output, we have two different tasks, i.e., answering single-answer questions and all questions. For single-answer questions, the models need to perform the single-label classification and output a score vector $\mathit{score}^\mathit{single}\in\mathbb{R}^n$ for each question, denoting the probability of each option being correct. For all questions, the models need to output a score vector $\mathit{score}^\mathit{all}$ of length $2^n-1$ for each question. Experimental results show that it's slightly better than using a score vector with legnth $n$. These values denote the probability of each possible combination of options.

Note that some models cannot be directly applied to our task, so we slightly modify them in the following steps:

\begin{figure}[t]
    \centering
    \includegraphics[width=.95\columnwidth]{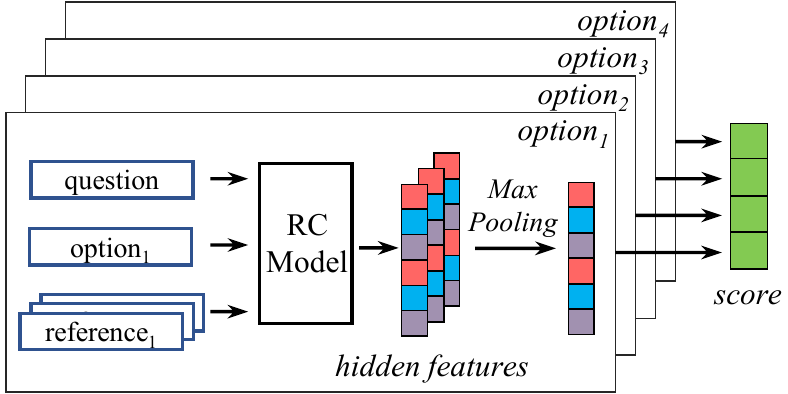}
    \caption{The unified framework for models on \dataset.}
    \label{fig:model}
\end{figure}

(1) If the original model only takes the questions and the reading paragraphs as input without options, we apply the model on the concatenation of the question and each option, and obtain a score $s_i$ for the $i$-th option. Then the score vector is represented as $\mathit{score}^\mathit{single}=[s_1,s_2,\ldots,s_n]$.

(2) If the original model is designed to extract answers from reading paragraphs, we modify the output layer into a linear layer that outputs the score of the $i$-th option, $s_i$.

(3) If the original model cannot be applied to multi-paragraph reading task, we apply the model on each reading paragraph of each option separately and the model will output the hidden representation $h_{i,j}\in\mathbb{R}^d$ for the $j$-th reading paragraph of the $i$-th option. We then employ max-pooling over all representations from the same option to obtain the hidden representation $h'_i$ for the $i$-th option that we have $h'_i=[h'_{i,1},h'_{i,2},\ldots,h'_{i,d}]$ where $h'_{i,j}=\max\left(h_{i,k,j}\mid \forall1\leq k\leq m\right)$. Finally, we pass $h'_i$ through a linear layer to obtain the score $s_i$ for the $i$-th option.
    
(4) We add a linear layer with input $\mathit{score}^\mathit{single}$ to obtain $\mathit{score}^\mathit{all}$ for answering all questions.

Besides, we adopt BertAdam~\cite{devlin2018bert} for Bert and Adam~\cite{kingma2014adam} for all other models. Meanwhile, for all experiments, we randomly select $20\%$ of the data as the test dataset. You can get more details from the website of the dataset.

\begin{table*}[htb]
\centering
\resizebox{1.6\columnwidth}{!}{
\begin{tabular}{c|cc|cc|cc}
\toprule
                                         & \multicolumn{2}{c|}{\typea} & \multicolumn{2}{c|}{\typeb} & \multicolumn{2}{c}{All} \\ \midrule
                                         & Single               & All                & Single               & All                & Single      & All       \\\midrule
Unskilled Humans        & $76.92$     & $71.11$     & $62.50$     & $58.00$     & $70.00$     & $64.21$     \\
Skilled Humans          & $80.64$     & $77.46$     & $86.84$     & $84.72$     & $84.06$     & $81.12$     \\
\midrule
\textbf{Co-matching}~\cite{wang2018co}     & $39.62$     & $\mathbf{25.37}$     & $\mathbf{48.91}$     & $28.61$     & $\mathbf{46.47}$     & $26.06$     \\
\textbf{BERT}~\cite{devlin2018bert}           & $38.05$     & $21.13$     & $38.89$     & $23.72$     & $39.56$     & $22.51$     \\
\textbf{SeaReader}~\cite{zhang2018medical}      & $39.29$     & $24.11$     & $45.32$     & $26.01$     & $40.50$     & $23.77$          \\
\textbf{Multi-Matching}~\cite{tang2019multi} & $\mathbf{41.96}$     & $23.63$     & $46.18$     & $\mathbf{29.06}$     & $42.98$     & $\mathbf{28.63}$       \\
\textbf{CSA}~\cite{chen2018convolutional}            & $32.44$     & -         & $34.76$     & -         & $21.03$     & -           \\
\textbf{CBM}~\cite{clark2017simple}         & $40.35$     & $22.54$     & $37.37$     & $22.50$     & $38.69$     & $22.53$          \\
\textbf{DSQA}~\cite{lin2018denoising}           & $34.15$     &      $18.41$    & $42.72$     &   $23.25$  & $42.63$     &     $22.69$  
\\ \bottomrule
\end{tabular}
}
\caption{Evaluation results (accuracy \%) of different models on \dataset.  Results marked ``-'' indicates that the model cannot converge within $256$ epochs.}
\label{table:result}
\end{table*}

\subsection{Baselines}

We implement $7$ representative reading comprehension and question answering models as our baselines, including:

\textbf{Co-matching}~\cite{wang2018co} achieves promising result on the RACE dataset~\cite{lai2017race}. The model matches reading paragraphs with questions and options with attention mechanism and uses the attention values to score options. This is a single-paragraph reading comprehension model for single-answer questions.

\textbf{BERT}~\cite{devlin2018bert} is the model which contains multiple bidirectional Transformer~\cite{vaswani2017attention} layers and has been fully pre-trained on large scaled datasets. As a single-paragraph reading comprehension model, \textbf{BERT} achieves state-of-the-art performance in most reading comprehension datasets including SQUAD~\cite{squaddataset}. We employ the base form of \textbf{BERT} pre-trained on Chinese documents in our experiments.

\textbf{SeaReader}~\cite{zhang2018medical} is proposed to answer questions in clinical medicine using knowledge extracted from publications in the medical domain. The model extracts information with question-centric attention, document-centric attention, and cross-document attention, and then uses a gated layer for denoising.

\textbf{Multi-Matching}~\cite{tang2019multi} employs Evidence-Answer Matching and Question-Passage-Answer Matching module to form matching information, and merges them together to obtain the scores of candidate answers.

\textbf{Convolutional Spatial Attention (CSA)}~\cite{chen2018convolutional} first generates enriched representations of passages, candidate answers, and questions with attention mechanism, and then applies CNN-MaxPooling operation to summarize adjacent attention information.

\textbf{Confidence-based Model (CBM)}~\cite{clark2017simple} is a simple and effective method for multi-paragraph reading comprehension task. They propose a pipeline method for single-paragraph reading comprehension and apply a confidence-based method to adapt the model to the multi-paragraph setting.

\textbf{Distantly Supervised Question Answering (DSQA)}~\cite{lin2018denoising} is an effective method for open-domain question answering, which decomposes the QA process into three steps: filter out noisy documents, extract correct answers and select the best answer.

\subsection{Experimental Results}

We evaluate the performance of all models on \dataset, with settings of single-answer question and all question answering. Besides, we also evaluate the performance in \typea\ and \typeb\ separately. In addition, we evaluate the performance of skilled and unskilled humans. Humans read the same paragraphs fetched by the search strategy as models do. Unskilled humans are those who do not have legal experience while skilled humans are those in legal professions. The experimental results are shown in Table~\ref{table:result}.

From these results, we observe that even the best-performed model can only achieve an accuracy of $28.63\%$ on all questions, while there is still a huge gap to $64\%$ accuracy for unskilled humans. We should note that unskilled humans read the same reading materials as models and they have no advanced knowledge about legal questions, so the gap mainly comes from the insufficiency of model reasoning ability. Meanwhile, compared with skilled humans, unskilled humans perform significantly worse than skilled humans on \typeb. The reason is that retrieved reading paragraphs are insufficient to provide enough evidence, as shown in Table~\ref{table:missing}, so the gap between unskilled and skilled humans mainly comes from the quality of retrieval.

Comparing the performance between \typea\ and \typeb, we reveal that most models achieve better performance on \typeb. Although a higher proportion of \typeb\ require multi-hop reasoning ability, the concepts in \typeb\ are always simpler ones, e.g., robbery, theft, or murder. The results also demonstrate that existing methods performs poorly in concept comprehension.

\subsection{Comparative Analysis}

We also perform a deeper analysis on the well-performed \textbf{Co-matching} by evaluating it on Annotation Set and the experimental results are listed in Table~\ref{table:reasonresults}. From the results we can see that existing methods can only answer about $32\%$ of questions correctly even when there is enough evidence in reading paragraphs, which means that the models cannot understand the reading materials at all. Moreover, we can see that the model performs extremely bad on multi-paragraph reading and multi-hop reasoning questions of \typeb. It means existing models cannot do multi-paragraph reading and multi-hop reasoning on real cases properly.

\begin{table}[tb]
\centering
\resizebox{.95\columnwidth}{!}{
\begin{tabular}{c|c|c|c}
\toprule
& KD-Q & CA-Q & All \\ \midrule
Word Matching               & $20.20$     & $28.00$     & $31.91$     \\
Concept Understanding       & $30.35$     & $20.83$     & $28.24$      \\
Numerical Analysis          & $16.67$     & $25.71$     & $30.00$     \\
Multi-Paragraph Reading     & $23.33$     & $19.44$     & $30.51$    \\
Multi-Hop Reasoning         & $25.00$     & $18.62$     & $30.30$        \\ \midrule
All Hit                     & $22.34$     & $24.47$     & $31.71$      \\
Partial Miss                & $29.73$     & $24.00$     & $26.76$       \\
All Miss                    & $21.05$     & $16.36$     & $29.79$      \\ \bottomrule
\end{tabular}
}
\caption{Performance of Co-matching on different questions.}
\label{table:reasonresults}
\end{table}

Besides, we also perform experiments with different value of $K$ on single \typea, and the experimental results are shown in Table~\ref{table:kresult}. From the results, we can see that more reading paragraphs cannot help the models to answer the questions better, as important articles have already been fetched even $K$ is small. It proves that the bad performance of models is because the insufficiency of reasoning ability rather than the quality of retrieval. As a larger value of $K$ cannot help with the accuracy, we select $K=6$ to reach a balance of speed and performance.

\begin{table}[htb]
\centering
\resizebox{.95\columnwidth}{!}{
\begin{tabular}{c|c|c|c|c|c|c}
\toprule
$K=$ & $1$ & $3$ & $6$ & $12$ & $18$ & $24$\\ \midrule
Accuracy & $30.1$ & $37.9$ & $39.6$ & $40.7$ & $40.7$  & $40.7$
\\\bottomrule
\end{tabular}
}
\caption{Performance of Co-matching with different $K$.}
\label{table:kresult}
\end{table}
 
\subsection{Case Study}

As shown in Table~\ref{table:exampleF}, we select an example to give an intuitive illustration on dealing with multi-hop reasoning. For most reading comprehension models, they choose all the options as their answers. Even without reading the statement, we can find that the option D conflicts with the other three options. Existing methods cannot handle conflicting options. Moreover, if we ignore option D, these models still choose all the remaining options, while the correct answer only contains option C. The models can easily find the evidence of option A, B, C from the statement with one-hop reasoning. However, if we read the related paragraphs, we will find the fact that Bob is under the age of $16$, which will filter out the options A and B. We can learn that existing reading comprehension models already have the ability of one-hop reasoning, but multi-hop reasoning is still challenging for them.

\begin{table}[h]
\centering
\resizebox{.95\columnwidth}{!}{
\begin{tabular}{p{\columnwidth}}
\toprule
\textbf{Question:} 
Bob is a male \textbf{born on February $\mathbf{27}$, $\mathbf{1987}$}.
Bob \emph{stole} from Alice a total of $5,000$ yuan in cash, one laptop (worth $13,000$ yuan), and other small jewelry on \textbf{February $\mathbf{27}$, $\mathbf{2003}$}.
While Bob was climbing back over the wall, Bob was seen by Catherine.
To escape, Bob quickly took a dagger from his pocket and stabbed in Catherine's heart, \emph{killing Catherine}.
So how should Bob's behavior be handled?\\\midrule
\textbf{Options:}\\
$\times$ (A). The crime of robbery.\\
$\times$ (B). The crime of theft.\\
$\checkmark$ (C). The ground of intentional homicide.\\
$\times$ (D). Bob does not constitute a crime.\\\midrule
\textbf{Paragraphs:}\\
\textbf{1.} A person who has \textbf{reached the age of $\mathbf{14}$ and under 16} will not constitute the crime of robbery and theft.\\
\textbf{2.} Calculation of age. ... For example, you are $\mathbf{14}$ years old from \textbf{the next day after your $\mathbf{14}$-th birthday}.\\
\bottomrule
\end{tabular}
}
\caption{A multi-hop reasoning example.}
\label{table:exampleF}
\end{table}
\section{Conclusion}

In this work, we present \dataset\ as a new and challenging dataset for LQA, and \dataset\ is the largest dataset in LQA. Both retrieving documents and answering questions in \dataset\ require multiple types of reasoning ability, and our experimental results show that existing state-of-the-art models cannot perform well on \dataset. We hope our \dataset\ can benefit researchers on improving the reasoning ability of reading comprehension and QA models, and also making advances for legal question answering.

In the future, we will explore how to improve the reasoning ability of question answering model and integrate legal knowledge into question answering, which are necessary for answering questions in \dataset.

\section{Acknowledgements}

This work is supported by the National Key Research and Development Program of China (No. 2018YFC0831900) and the National Natural Science Foundation of China (NSFC No. 61572273, 61661146007).

\small{
\bibliography{AAAI-ZhongH.5727}

\begin{thebibliography}{}

\bibitem[\protect\citeauthoryear{Berant \bgroup et al\mbox.\egroup
  }{2013}]{berant2013semantic}
Berant, J.; Chou, A.; Frostig, R.; and Liang, P.
\newblock 2013.
\newblock Semantic parsing on freebase from question-answer pairs.
\newblock In {\em Proceedings of EMNLP}.

\bibitem[\protect\citeauthoryear{Bordes \bgroup et al\mbox.\egroup
  }{2015}]{bordes2015large}
Bordes, A.; Usunier, N.; Chopra, S.; and Weston, J.
\newblock 2015.
\newblock Large-scale simple question answering with memory networks.
\newblock {\em arXiv preprint arXiv:1506.02075}.

\bibitem[\protect\citeauthoryear{Chen and
  Van~Durme}{2017}]{chen2017discriminative}
Chen, T., and Van~Durme, B.
\newblock 2017.
\newblock Discriminative information retrieval for question answering sentence
  selection.
\newblock In {\em Proceedings of EACL}.

\bibitem[\protect\citeauthoryear{Chen \bgroup et al\mbox.\egroup
  }{2017}]{chen2017reading}
Chen, D.; Fisch, A.; Weston, J.; and Bordes, A.
\newblock 2017.
\newblock Reading wikipedia to answer open-domain questions.
\newblock In {\em Proceedings of ACL}.

\bibitem[\protect\citeauthoryear{Chen \bgroup et al\mbox.\egroup
  }{2019}]{chen2018convolutional}
Chen, Z.; Cui, Y.; Ma, W.; Wang, S.; and Hu, G.
\newblock 2019.
\newblock Convolutional spatial attention model for reading comprehension with
  multiple-choice questions.
\newblock In {\em Proceedings of AAAI}.

\bibitem[\protect\citeauthoryear{Chen, Liu, and Ho}{2013}]{chen2013text}
Chen, Y.-L.; Liu, Y.-H.; and Ho, W.-L.
\newblock 2013.
\newblock A text mining approach to assist the general public in the retrieval
  of legal documents.
\newblock {\em Journal of ASIS\&T} 64(2):280--290.

\bibitem[\protect\citeauthoryear{Clark and Gardner}{2018}]{clark2017simple}
Clark, C., and Gardner, M.
\newblock 2018.
\newblock Simple and effective multi-paragraph reading comprehension.
\newblock In {\em Proceedings of ACL}.

\bibitem[\protect\citeauthoryear{Clark \bgroup et al\mbox.\egroup
  }{2018}]{clark2018think}
Clark, P.; Cowhey, I.; Etzioni, O.; Khot, T.; Sabharwal, A.; Schoenick, C.; and
  Tafjord, O.
\newblock 2018.
\newblock Think you have solved question answering? try arc, the ai2 reasoning
  challenge.
\newblock {\em arXiv preprint arXiv:1803.05457}.

\bibitem[\protect\citeauthoryear{Cui \bgroup et al\mbox.\egroup
  }{2017}]{cui2016attention}
Cui, Y.; Chen, Z.; Wei, S.; Wang, S.; Liu, T.; and Hu, G.
\newblock 2017.
\newblock Attention-over-attention neural networks for reading comprehension.
\newblock In {\em Proceedings of ACL}.

\bibitem[\protect\citeauthoryear{Devlin \bgroup et al\mbox.\egroup
  }{2018}]{devlin2018bert}
Devlin, J.; Chang, M.-W.; Lee, K.; and Toutanova, K.
\newblock 2018.
\newblock Bert: Pre-training of deep bidirectional transformers for language
  understanding.
\newblock {\em arXiv preprint arXiv:1810.04805}.

\bibitem[\protect\citeauthoryear{Dhingra \bgroup et al\mbox.\egroup
  }{2017}]{dhingra2016gated}
Dhingra, B.; Liu, H.; Yang, Z.; Cohen, W.~W.; and Salakhutdinov, R.
\newblock 2017.
\newblock Gated-attention readers for text comprehension.
\newblock In {\em Proceedings of ACL}.

\bibitem[\protect\citeauthoryear{Do \bgroup et al\mbox.\egroup
  }{2017}]{do2017legal}
Do, P.-K.; Nguyen, H.-T.; Tran, C.-X.; Nguyen, M.-T.; and Nguyen, M.-L.
\newblock 2017.
\newblock Legal question answering using ranking svm and deep convolutional
  neural network.
\newblock {\em arXiv preprint arXiv:1703.05320}.

\bibitem[\protect\citeauthoryear{Fawei \bgroup et al\mbox.\egroup
  }{2018}]{fawei2018methodology}
Fawei, B.; Pan, J.~Z.; Kollingbaum, M.; and Wyner, A.~Z.
\newblock 2018.
\newblock A methodology for a criminal law and procedure ontology for legal
  question answering.
\newblock In {\em Proceedings of JIST}.

\bibitem[\protect\citeauthoryear{Green~Jr \bgroup et al\mbox.\egroup
  }{1961}]{green1961baseball}
Green~Jr, B.~F.; Wolf, A.~K.; Chomsky, C.; and Laughery, K.
\newblock 1961.
\newblock Baseball: an automatic question-answerer.
\newblock In {\em Proceedings of IRE-AIEE-ACM}.

\bibitem[\protect\citeauthoryear{He \bgroup et al\mbox.\egroup
  }{2018a}]{he2018secaps}
He, C.; Peng, L.; Le, Y.; and He, J.
\newblock 2018a.
\newblock Secaps: A sequence enhanced capsule model for charge prediction.
\newblock {\em arXiv preprint arXiv:1810.04465}.

\bibitem[\protect\citeauthoryear{He \bgroup et al\mbox.\egroup
  }{2018b}]{dureader2017he}
He, W.; Liu, K.; Liu, J.; Lyu, Y.; Zhao, S.; Xiao, X.; Liu, Y.; Wang, Y.; Wu,
  H.; She, Q.; et~al.
\newblock 2018b.
\newblock Dureader: a chinese machine reading comprehension dataset from
  real-world applications.
\newblock In {\em Proceedings of ACL workshop}.

\bibitem[\protect\citeauthoryear{Hermann \bgroup et al\mbox.\egroup
  }{2015}]{cnndailymaildataset}
Hermann, K.~M.; Kocisky, T.; Grefenstette, E.; Espeholt, L.; Kay, W.; Suleyman,
  M.; and Blunsom, P.
\newblock 2015.
\newblock Teaching machines to read and comprehend.
\newblock In {\em Proceedings of NIPS}.

\bibitem[\protect\citeauthoryear{Hu \bgroup et al\mbox.\egroup
  }{2018}]{hu2018few}
Hu, Z.; Li, X.; Tu, C.; Liu, Z.; and Sun, M.
\newblock 2018.
\newblock Few-shot charge prediction with discriminative legal attributes.
\newblock In {\em Proceedings of COLING}.

\bibitem[\protect\citeauthoryear{Jia and Liang}{2017}]{jia2017adversarial}
Jia, R., and Liang, P.
\newblock 2017.
\newblock Adversarial examples for evaluating reading comprehension systems.
\newblock In {\em Proceedings of EMNLP}.

\bibitem[\protect\citeauthoryear{Johnson and Zhang}{2017}]{johnson2017deep}
Johnson, R., and Zhang, T.
\newblock 2017.
\newblock Deep pyramid convolutional neural networks for text categorization.
\newblock In {\em Proceedings of ACL}.

\bibitem[\protect\citeauthoryear{Joshi \bgroup et al\mbox.\egroup
  }{2017}]{triviaqa2017}
Joshi, M.; Choi, E.; Weld, D.~S.; and Zettlemoyer, L.
\newblock 2017.
\newblock Triviaqa: A large scale distantly supervised challenge dataset for
  reading comprehension.
\newblock {\em arXiv preprint arXiv:1705.03551}.

\bibitem[\protect\citeauthoryear{Joulin \bgroup et al\mbox.\egroup
  }{2017}]{joulin2016fasttext}
Joulin, A.; Grave, E.; Bojanowski, P.; Douze, M.; J{\'e}gou, H.; and Mikolov,
  T.
\newblock 2017.
\newblock Fasttext. zip: Compressing text classification models.
\newblock In {\em Proceedings of ICLR}.

\bibitem[\protect\citeauthoryear{Kim \bgroup et al\mbox.\egroup
  }{2016}]{kim2016coliee}
Kim, M.-Y.; Goebel, R.; Kano, Y.; and Satoh, K.
\newblock 2016.
\newblock Coliee-2016: evaluation of the competition on legal information
  extraction and entailment.
\newblock In {\em Proceedings of JURISIN}.

\bibitem[\protect\citeauthoryear{Kim \bgroup et al\mbox.\egroup
  }{2018}]{kim2018coliee}
Kim, M.-Y.; Lu, Y.; Rabelo, J.; and Goebel, R.
\newblock 2018.
\newblock Coliee-2018: Evaluation of the competition on case law information
  extraction and entailment.

\bibitem[\protect\citeauthoryear{Kim}{2014}]{kim2014convolutional}
Kim, Y.
\newblock 2014.
\newblock Convolutional neural networks for sentence classification.
\newblock In {\em Proceedings of EMNLP}.

\bibitem[\protect\citeauthoryear{Kingma and Ba}{2015}]{kingma2014adam}
Kingma, D., and Ba, J.
\newblock 2015.
\newblock Adam: A method for stochastic optimization.
\newblock In {\em Proceedings of ICLR}.

\bibitem[\protect\citeauthoryear{Kwok, Etzioni, and
  Weld}{2001}]{kwok2001scaling}
Kwok, C.; Etzioni, O.; and Weld, D.~S.
\newblock 2001.
\newblock Scaling question answering to the web.
\newblock {\em ACM Transactions on Information Systems}.

\bibitem[\protect\citeauthoryear{Lai \bgroup et al\mbox.\egroup
  }{2017}]{lai2017race}
Lai, G.; Xie, Q.; Liu, H.; Yang, Y.; and Hovy, E.
\newblock 2017.
\newblock {RACE}: Large-scale reading comprehension dataset from examinations.
\newblock In {\em Proceedings of EMNLP}.

\bibitem[\protect\citeauthoryear{Lin \bgroup et al\mbox.\egroup
  }{2018}]{lin2018denoising}
Lin, Y.; Ji, H.; Liu, Z.; and Sun, M.
\newblock 2018.
\newblock Denoising distantly supervised open-domain question answering.
\newblock In {\em Proceedings of ACL}.

\bibitem[\protect\citeauthoryear{Luo \bgroup et al\mbox.\egroup
  }{2017}]{luo2017learning}
Luo, B.; Feng, Y.; Xu, J.; Zhang, X.; and Zhao, D.
\newblock 2017.
\newblock Learning to predict charges for criminal cases with legal basis.
\newblock In {\em Proceedings of EMNLP}.

\bibitem[\protect\citeauthoryear{Nguyen \bgroup et al\mbox.\egroup
  }{2016}]{msmarco2016}
Nguyen, T.; Rosenberg, M.; Song, X.; Gao, J.; Tiwary, S.; Majumder, R.; and
  Deng, L.
\newblock 2016.
\newblock Ms marco: A human generated machine reading comprehension dataset.
\newblock {\em arXiv preprint arXiv:1611.09268}.

\bibitem[\protect\citeauthoryear{Raghav, Reddy, and
  Reddy}{2016}]{raghav2016analyzing}
Raghav, K.; Reddy, P.~K.; and Reddy, V.~B.
\newblock 2016.
\newblock Analyzing the extraction of relevant legal judgments using
  paragraph-level and citation information.
\newblock In {\em Proceedings of ECAI}.

\bibitem[\protect\citeauthoryear{Rajpurkar \bgroup et al\mbox.\egroup
  }{2016}]{squaddataset}
Rajpurkar, P.; Zhang, J.; Lopyrev, K.; and Liang, P.
\newblock 2016.
\newblock Squad: 100,000+ questions for machine comprehension of text.
\newblock In {\em Proceedings of EMNLP}.

\bibitem[\protect\citeauthoryear{Rajpurkar, Jia, and
  Liang}{2018}]{rajpurkar2018know}
Rajpurkar, P.; Jia, R.; and Liang, P.
\newblock 2018.
\newblock Know what you don't know: Unanswerable questions for squad.
\newblock In {\em Proceedings of ACL}.

\bibitem[\protect\citeauthoryear{Richardson, Burges, and
  Renshaw}{2013}]{mctestdataset}
Richardson, M.; Burges, C.~J.; and Renshaw, E.
\newblock 2013.
\newblock Mctest: A challenge dataset for the open-domain machine comprehension
  of text.
\newblock In {\em Proceedings of EMNLP}.

\bibitem[\protect\citeauthoryear{Seo \bgroup et al\mbox.\egroup
  }{2017}]{seo2016bidirectional}
Seo, M.; Kembhavi, A.; Farhadi, A.; and Hajishirzi, H.
\newblock 2017.
\newblock Bidirectional attention flow for machine comprehension.
\newblock In {\em Proceedings of ICLR}.

\bibitem[\protect\citeauthoryear{Shen \bgroup et al\mbox.\egroup
  }{2018}]{shen2018legal}
Shen, Y.; Sun, J.; Li, X.; Zhang, L.; Li, Y.; and Shen, X.
\newblock 2018.
\newblock Legal article-aware end-to-end memory network for charge prediction.
\newblock In {\em Proceedings of ICCSE}.

\bibitem[\protect\citeauthoryear{Tang, Cai, and Zhuo}{2019}]{tang2019multi}
Tang, M.; Cai, J.; and Zhuo, H.~H.
\newblock 2019.
\newblock Multi-matching network for multiple choice reading comprehension.
\newblock In {\em Proceedings of AAAI}.

\bibitem[\protect\citeauthoryear{Trischler \bgroup et al\mbox.\egroup
  }{2016}]{newsqa2016}
Trischler, A.; Wang, T.; Yuan, X.; Harris, J.; Sordoni, A.; Bachman, P.; and
  Suleman, K.
\newblock 2016.
\newblock Newsqa: A machine comprehension dataset.
\newblock {\em arXiv preprint arXiv:1611.09830}.

\bibitem[\protect\citeauthoryear{Vaswani \bgroup et al\mbox.\egroup
  }{2017}]{vaswani2017attention}
Vaswani, A.; Shazeer, N.; Parmar, N.; Uszkoreit, J.; Jones, L.; Gomez, A.~N.;
  Kaiser, {\L}.; and Polosukhin, I.
\newblock 2017.
\newblock Attention is all you need.
\newblock In {\em Proceedings of NIPS}.

\bibitem[\protect\citeauthoryear{Voorhees and others}{1999}]{voorhees1999trec}
Voorhees, E.~M., et~al.
\newblock 1999.
\newblock The trec-8 question answering track report.
\newblock In {\em Proceedings of Trec}.

\bibitem[\protect\citeauthoryear{Wang and Jiang}{2016}]{wang2016RC}
Wang, S., and Jiang, J.
\newblock 2016.
\newblock Machine comprehension using match-lstm and answer pointer.
\newblock {\em arXiv preprint arXiv:1608.07905}.

\bibitem[\protect\citeauthoryear{Wang \bgroup et al\mbox.\egroup
  }{2017}]{wang2017gated}
Wang, W.; Yang, N.; Wei, F.; Chang, B.; and Zhou, M.
\newblock 2017.
\newblock Gated self-matching networks for reading comprehension and question
  answering.
\newblock In {\em Proceedings of ACL}.

\bibitem[\protect\citeauthoryear{Wang \bgroup et al\mbox.\egroup
  }{2018a}]{wang2018co}
Wang, S.; Yu, M.; Chang, S.; and Jiang, J.
\newblock 2018a.
\newblock A co-matching model for multi-choice reading comprehension.
\newblock In {\em Proceedings of ACL}.

\bibitem[\protect\citeauthoryear{Wang \bgroup et al\mbox.\egroup
  }{2018b}]{wang2017r}
Wang, S.; Yu, M.; Guo, X.; Wang, Z.; Klinger, T.; Zhang, W.; Chang, S.;
  Tesauro, G.; Zhou, B.; and Jiang, J.
\newblock 2018b.
\newblock R$^{3}$: Reinforced reader-ranker for open-domain question answering.
\newblock In {\em Proceedings of AAAI}.

\bibitem[\protect\citeauthoryear{Wang \bgroup et al\mbox.\egroup
  }{2018c}]{wang2017evidence}
Wang, S.; Yu, M.; Jiang, J.; Zhang, W.; Guo, X.; Chang, S.; Wang, Z.; Klinger,
  T.; Tesauro, G.; and Campbell, M.
\newblock 2018c.
\newblock Evidence aggregation for answer re-ranking in open-domain question
  answering.
\newblock In {\em Proceedings of ICLR}.

\bibitem[\protect\citeauthoryear{Wang \bgroup et al\mbox.\egroup
  }{2018d}]{wang2018multi}
Wang, Y.; Liu, K.; Liu, J.; He, W.; Lyu, Y.; Wu, H.; Li, S.; and Wang, H.
\newblock 2018d.
\newblock Multi-passage machine reading comprehension with cross-passage answer
  verification.
\newblock In {\em Proceedings of ACL}.

\bibitem[\protect\citeauthoryear{Xiao \bgroup et al\mbox.\egroup
  }{2018}]{xiao2018cail}
Xiao, C.; Zhong, H.; Guo, Z.; Tu, C.; Liu, Z.; Sun, M.; Feng, Y.; Han, X.; Hu,
  Z.; Wang, H.; et~al.
\newblock 2018.
\newblock Cail2018: A large-scale legal dataset for judgment prediction.
\newblock {\em arXiv preprint arXiv:1807.02478}.

\bibitem[\protect\citeauthoryear{Yang \bgroup et al\mbox.\egroup
  }{2018}]{yang2018hotpotqa}
Yang, Z.; Qi, P.; Zhang, S.; Bengio, Y.; Cohen, W.; Salakhutdinov, R.; and
  Manning, C.~D.
\newblock 2018.
\newblock Hotpotqa: A dataset for diverse, explainable multi-hop question
  answering.
\newblock In {\em Proceedings of EMNLP}.

\bibitem[\protect\citeauthoryear{Yang, Yih, and Meek}{2015}]{yang2015wikiqa}
Yang, Y.; Yih, W.-t.; and Meek, C.
\newblock 2015.
\newblock Wikiqa: A challenge dataset for open-domain question answering.
\newblock In {\em Proceedings of EMNLP}.

\bibitem[\protect\citeauthoryear{Ye \bgroup et al\mbox.\egroup
  }{2018}]{ye2018interpretable}
Ye, H.; Jiang, X.; Luo, Z.; and Chao, W.
\newblock 2018.
\newblock Interpretable charge predictions for criminal cases: Learning to
  generate court views from fact descriptions.
\newblock In {\em Proceedings of NAACL}.

\bibitem[\protect\citeauthoryear{Yih \bgroup et al\mbox.\egroup
  }{2015}]{yih2015semantic}
Yih, W.-t.; Chang, M.-W.; He, X.; and Gao, J.
\newblock 2015.
\newblock Semantic parsing via staged query graph generation: Question
  answering with knowledge base.
\newblock In {\em Proceedings of ACL}.

\bibitem[\protect\citeauthoryear{Yu \bgroup et al\mbox.\egroup
  }{2017}]{yu2017improved}
Yu, M.; Yin, W.; Hasan, K.~S.; Santos, C.~d.; Xiang, B.; and Zhou, B.
\newblock 2017.
\newblock Improved neural relation detection for knowledge base question
  answering.
\newblock In {\em Proceedings of ACL}.

\bibitem[\protect\citeauthoryear{Zhang \bgroup et al\mbox.\egroup
  }{2018}]{zhang2018medical}
Zhang, X.; Wu, J.; He, Z.; Liu, X.; and Su, Y.
\newblock 2018.
\newblock Medical exam question answering with large-scale reading
  comprehension.
\newblock In {\em Proceedings of AAAI}.

\bibitem[\protect\citeauthoryear{Zhong \bgroup et al\mbox.\egroup
  }{2018}]{zhong2018legal}
Zhong, H.; Zhipeng, G.; Tu, C.; Xiao, C.; Liu, Z.; and Sun, M.
\newblock 2018.
\newblock Legal judgment prediction via topological learning.
\newblock In {\em Proceedings of EMNLP}.

\end{thebibliography}
\bibliographystyle{aaai}
}

\end{document}